\documentclass{bmvc2k}

\usepackage{appendix}
\usepackage{amsfonts}
\usepackage{siunitx}
\usepackage{multicol}
\usepackage{booktabs}
\title{clip2latent: Text driven sampling of a pre-trained StyleGAN using denoising diffusion and CLIP}

\addauthor{Justin N. M. Pinkney}{justin@lambdal.com}{1}
\addauthor{Chuan Li}{c@lambdal.com}{1}

\addinstitution{
 Lambda, Inc\\
 San Francisco, USA
}

\runninghead{J.N.M. Pinkney, C. Li}{clip2latent}

\begin{document}
\maketitle

\begin{abstract}
We introduce a new method to efficiently create text-to-image models from a pre-trained CLIP and StyleGAN. It enables text driven sampling with an existing generative model without any external data or fine-tuning. This is achieved by training a diffusion model conditioned on CLIP embeddings to sample latent vectors of a pre-trained StyleGAN, which we call \textit{clip2latent}. We leverage the alignment between CLIP’s image and text embeddings to avoid the need for any text labelled data for training the conditional diffusion model. We demonstrate that clip2latent allows us to generate high-resolution (1024x1024 pixels) images based on text prompts with fast sampling, high image quality, and low training compute and data requirements. We also show that the use of the well studied StyleGAN architecture, without further fine-tuning, allows us to directly apply existing methods to control and modify the generated images adding a further layer of control to our text-to-image pipeline.

\end{abstract}

\begin{figure}[htbp]
    \centering
    \includegraphics[width=0.955\textwidth]{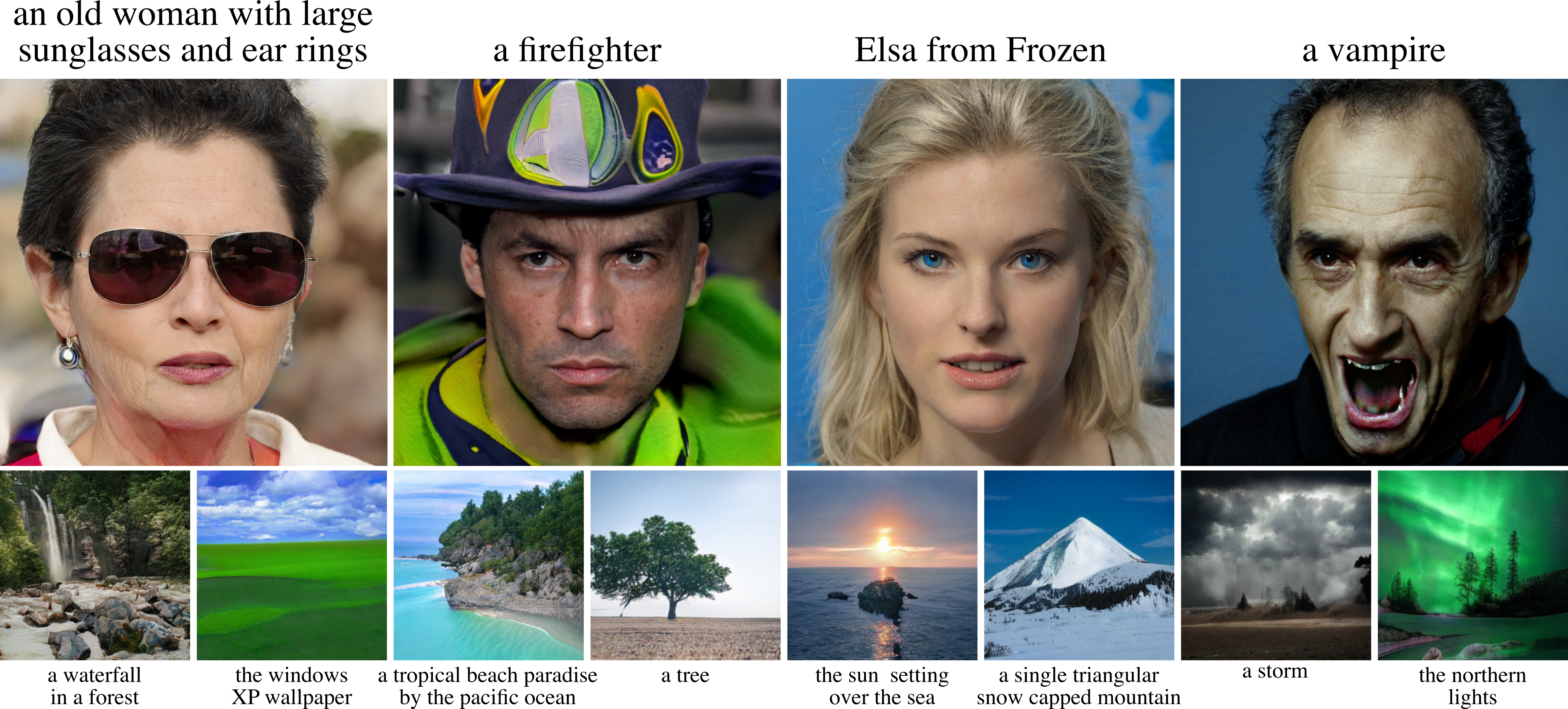}
    \vspace{-0.1cm}
    \caption{Images generated from text prompts by clip2latent trained on: Top: 1024x1024 StyleGAN2 FFHQ model. Bottom: 256x256 StyleGAN3 LHQ model. Prompts are given next to images and all were prefixed with "A photograph of".}
    \label{fig:headline}
\end{figure}

\section{Introduction}

We take inspiration from the recent work of DALL-E 2\cite{Ramesh2022-ai} which trains a denoising diffusion model\cite{Song2020-cd,Ho2020-dk,Sohl-Dickstein2015-hz} to generate CLIP image embeddings from CLIP text embeddings. We generalise this approach to train a diffusion model to generate latent codes for a generative model conditional on CLIP image embeddings. 

In effect, we map between the latent spaces of two different pre-trained models to enable controllable generation from a previous unconditional generative model. In this case we choose to generate StyleGAN2/3\cite{Karras2019-sj,Karras2021-sg} latent codes conditional on CLIP\cite{Radford2021-ru} image/text embeddings to enable text conditional synthesis using the pre-trained StyleGAN model. As both the text encoder and image synthesiser are frozen this method is quick to train, and by leveraging the shared embedding space of images and text in CLIP we are able to train it with no external data.

At inference time our model generates high-quality and high-resolution outputs, well aligned to a text prompt, with minimal images artefacts. We also show that we can benefit from the hierarchical and controllable nature of StyleGAN's existing latent space which can further be used to exert control over the generated images, see Section \ref{sec:using}.

In summary, our main contributions are:
\begin{itemize}
  \item A conditional diffusion model that connects pre-trained CLIP and StyleGAN models into a text-to-image model.  
  \item A training scheme that requires no text-image pairs.
  \item A model which is quick to train, and can produce mega-pixel samples in under a second.
\end{itemize}

Source code and trained models used in this work are available at the following url: \url{https://github.com/justinpinkney/clip2latent}.

\section{Related work}

\subsection{Text-to-image generation}

Recent months have seen a tremendous advance in the quality of text-to-image generative models. The majority of these models are trained using large amounts of text-image paired data where, image generation is conditioned on the text input. To expedite the training and improve the quality of generation, several of the state of the art approaches make use of pre-trained models. Some use a pre-trained text encoder to leverage the semantic knowledge captured by large-scale text\cite{Saharia2022-ry} or multi-modal pre-training\cite{Ramesh2022-ai}. Others have used a pre-trained image auto-encoder to reduce the effective spatial dimensionality required for acceptable resolution\cite{Rombach2021-bu,Gu2021-tt} and in some case to work with an image representation more amenable to the image synthesis model\cite{Ramesh2021-ro,Ding2022-gr}.  In contrast to our work however, the actual image synthesis model rarely makes use of an entirely pre-trained generator. 

Early work in leveraging only pre-trained models was inspired by feature visualisation approaches\cite{Olah2017-lu}, applying a variety of image parameterisations and regularisation allowed a successful iterative optimisation to maximise the similarity of an image to a given text prompt\cite{murdock1}. As well as conventional image parameterisations, this approach has also been applied to generative models where the latent code is directly optimised to produce model outputs which best match a given text prompt\cite{murdock2}. This approach is conceptually closest to ours, where we replace the time consuming iterative optimisation with sampling from a conditional diffusion model which generates latent codes. 

Another notable approach which leverages only pre-trained models is CLIP guided diffusion, where the classifier guidance approach for sampling a diffusion model can be applied using CLIP as the classifier\cite{crowson1}. However, this classifier guidance approach has since been shown to be limited by the lack of pre-trained CLIP models trained on noised images and both approaches are outperformed by classifier-free guidance for text-to-image generation\cite{Nichol2021-be}.

\subsection{Text driven manipulation of pre-trained StyleGAN}

Although there has been limited work in using a pre-trained StyleGAN purely for text-to-image generation, there has been exploration of text driven \textit{editing} using StyleGAN models. StyleCLIP\cite{Patashnik2021-zm}, CLIP2StyleGAN\cite{Abdal2021-lv}, StyleMC\cite{Kocasari2021-wv}, and others\cite{Xu2021-ev,Gabbay2021-go,Tzelepis2022-fj} use a variety of methods to extract editing directions in the latent space of a pre-trained StyleGAN based on text descriptions via CLIP. Although not the main application of our work, we note that our approach can also be applied to discovery of edit directions and show a small proof-of-concept example in Appendix F.

Finally, StyleGAN-Nada\cite{Gal2021-ir} performs domain adaptation of a pre-trained StyleGAN model by fine-tuning the generator based on a text description of the desired domain encoded by CLIP. Mind the Gap\cite{Zhu2021-bw} takes a similar approach, but uses CLIP image encodings to perform domain adaptation to a new visual style. Unlike our approach this requires fine-tuning of the generator which modifies its entire output to match a single text/image prompt without adding any further text based control.

\subsection{Language-Free training}

One of our key contributions is the ability to train an text-to-image model without any text labelled training data. Two previous works have shown similar results. LAFITE\cite{Zhou2021-dl} trains a GAN conditioned on CLIP image embeddings, we compare our results to their language-free text-to-image model in Section \ref{sec:results}. Note the method of LAFITE requires training of the generator from scratch, whereas our method freezes the pre-trained StyleGAN generator. CLIP-GEN\cite{Wang2022-cf} similarly trains an auto-regressive transformer model to predict VQGAN tokens conditioned only on CLIP image embeddings and show that this approach generalises to CLIP text embeddings. In essence our method takes this, and similar\cite{Rombach2021-bu,Ramesh2021-ro}, approaches to the extreme: predicting a single StyleGAN latent vector rather than VQGAN tokens. This enables our model to be far smaller (48M vs 307M-1.6B parameters) and train much faster (500k vs 2M iterations) in exchange for being constrained closely to the original domain of the generative model.

\section{Approach}

Our approach is to map the latent spaces of pre-trained CLIP and StyleGAN models using a diffusion model we call clip2latent. 

\begin{figure}[htbp]
    \centering
    \includegraphics[width=\textwidth]{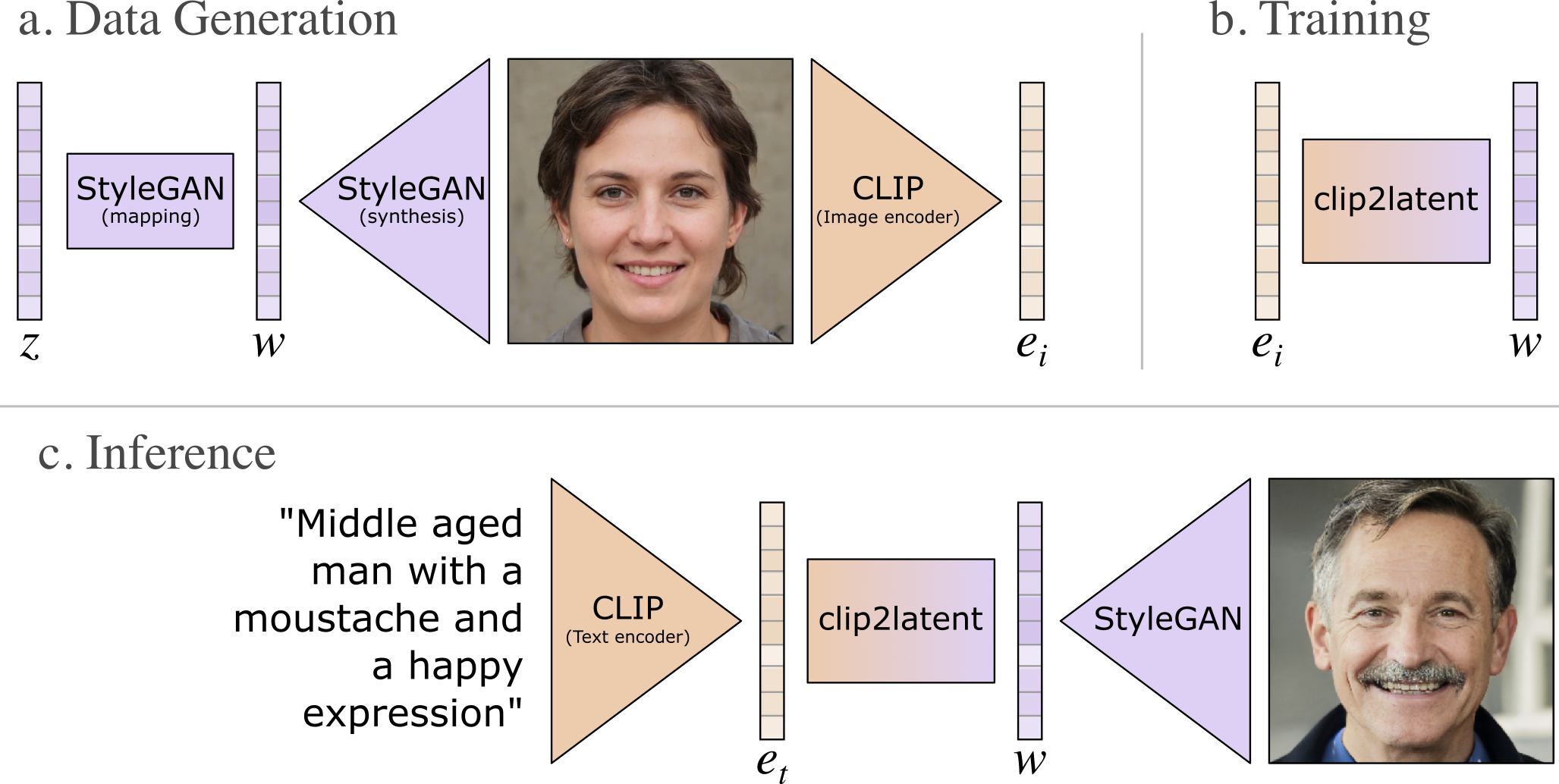}
    \caption{A schematic diagram of: a. synthetic data generation b. clip2latent diffusion model training, and c. text-to-image inference. Triangle and rectangle blocks indicate models, vertical striped bars indicate latent vectors or embeddings. $z$ and $w$ are the StyleGAN $z$ and $w$ latent spaces, $e_i$ and $e_t$ denote the CLIP embeddings space for images and text respectively.}
    \label{fig:training}
\end{figure}

StyleGAN is a powerful generative model which can be sampled with a normally distributed latent vector $z \in\mathbb{R}^{512}$. From this vector StyleGAN applies a 'mapping network' to map to a second latent space, $w \in\mathbb{R}^{512}$, finally this $w$ vector is used to generate an image. To generate our dataset we randomly sample $z$ latents and generate images from these with StyleGAN. Next we use the image encoder component of CLIP to encode StyleGAN generated images into the CLIP embedding space $e_i \in\mathbb{R}^{512}$. After encoding the StyleGAN generated images using CLIP, we can discard the generated images and original $z$ latents to leave our training data $(w, e_i)$ tuples, see Figure \ref{fig:training}a.

We then train the clip2latent model using the same approach as the diffusion based prior from DALL-E 2\cite{Ramesh2022-ai}, Figure \ref{fig:training}b. Briefly we train a Gaussian diffusion model to generate $w$ vectors conditioned on $e_i$ embeddings. As in DALL-E 2 we train a transformer based model on a sequence consisting of: the CLIP image embedding $e_i$, an embedding representing the timestep in the diffusion process, the noised $w$ latent vector, and a final learned embedding whos output predicts the denoised $w$.

Once we have trained clip2latent, we rely on the fact that CLIP can embed images and text into a shared latent space. At inference time we generate a CLIP embedding for a text description ($e_t$) and use this as the conditioning to generate a StyleGAN latent vector, from which we can create a high-resolution image using the StyleGAN generator, illustrated in Figure \ref{fig:training}c.

\section{Training details}

\subsection{Data generation}

To generate paired training samples without the need for external images and text labels we use the pre-trained StyleGAN 2 model (config-F) trained on FFHQ\cite{Karras2019-sj}. We sample 1 million random latent vectors with no truncation and generate the CLIP embeddings for each image using the ViT-B/32 pre-trained CLIP\cite{Radford2021-ru}. We resize the generated StyleGAN image to 224x224 pixels and apply the CLIP image normalisation. For each image we generate a single embedding and do not perform any data augmentation on the generated images.

\subsection{Latent prior training}

We then train a denoising diffusion model to generate the latent vectors conditioned on the CLIP image embeddings. We choose the same architecture as the prior in DALLE-2\cite{Ramesh2022-ai} and use the open source implementation DALL-E 2-Pytorch\cite{wang1}. Our diffusion model is analogous to that presented in DALLE-2 where the CLIP image embeddings are replaced by the StyleGAN latent codes and the CLIP text embedding are replaced by the corresponding CLIP image embeddings, as such we could consider this model to be a \emph{latent prior}.

During training we substitute the image embeddings with zeros with a probability of 0.2 to allow us to use classifier-free guidance\cite{Ho2021-ck} during inference. Rather than directly use the $w$ latent vectors for training we subtract the mean and divide by the estimated standard deviation of samples from $w$ space to better match the variance expected by the diffusion model. This also aligns with the common practice of learning StyleGAN latent prediction models relative to the mean latent vector\cite{Richardson2020-og}.

To improve the ability of the model to generalise to text embeddings and avoid over-fitting on image embeddings we take the approach introduced in LAFITE\cite{Zhou2021-dl} where "pseudo-text" embeddings are generated from image embeddings using scaled Gaussian noise, see Section \ref{sec:noise} and Appendix A.1.

We train our diffusion model for 1 million iterations at a batch size of 512, for full hyperparameter details see Appendix A. We select the best checkpoint as judged by generating 4 samples each for a set of 16 text prompts and computing the mean CLIP cosine similarity score between the text embedding and the CLIP image embedding of the generated image. In practice, we find that this peak validation score generally occurs at around 500 thousand iterations which takes approximately 9 hours on a single A100-80GB GPU.

\subsection{Text-to-image synthesis}

At inference time we generate the CLIP embedding for a text input and use this as the condition for the clip2latent network. By performing denoising diffusion, the clip2latent network generates a StyleGAN latent code which can then be used for image synthesis. During the denoising process we optionally employ timestep respacing\cite{Nichol2021-jd} and super conditioning\cite{Ho2021-ck} to decrease the sampling time and increase the CLIP similarity respectively. We do not employ any clipping or renormalisation of the denoised outputs or any of the recent method to reduce super-conditioned artefacts\cite{Karras2022-wf,Wang2022-ic}. In practice, with our shortest denoising schedule, end-to-end text to image synthesis can be performed in under a quarter of a second on a single A100 GPU for a single image (see Table \ref{table:compare}). As our image generation pipeline is fast and scales well to batching, we employ CLIP re-ranking to obtain the best CLIP similarity scoring sample from a batch of 16 candidates.

\section{Results}
\label{sec:results}

\begin{figure}[tbp]
    \centering
    \makebox[\textwidth][c]{\includegraphics[width=\textwidth]{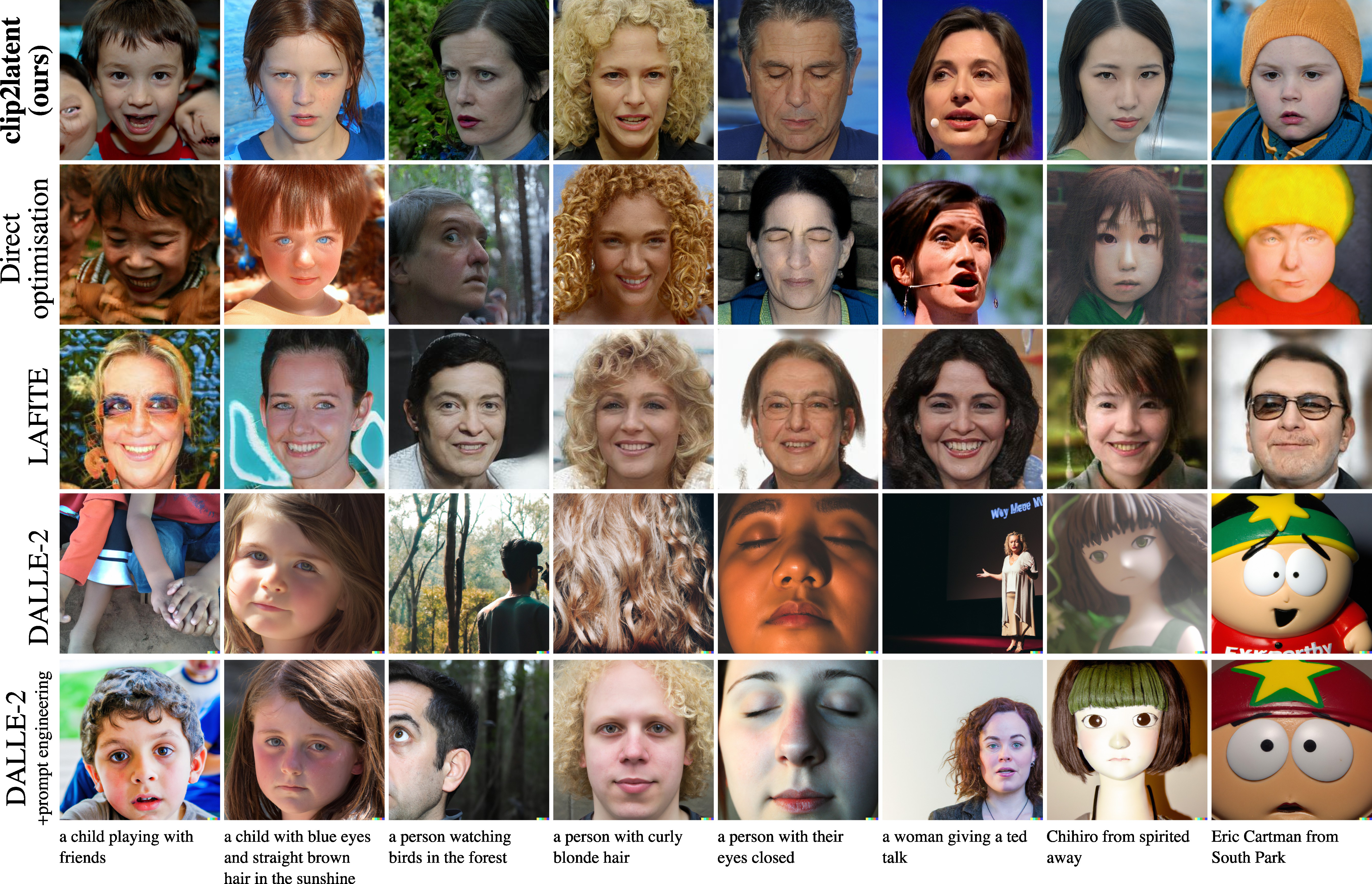}}
    \caption{Sample generations from our model (clip2latent) compared to direct optimisation of the StyleGAN latent vectors, LAFITE, and DALLE-2. For details of the comparison methods see Appendix C. Note, LAFITE generates samples at 256x256 pixels, whereas clip2latent and direct optimisation generate samples at 1024x1024. All prompts were prefixed with "A photograph of".}
    \label{fig:compare}
\end{figure}

\renewcommand{\arraystretch}{1.2}
\begin{table}

    \centering
    \caption{CLIP similarity scores and inference times for our methods compared to previous work. Similarity scores are measured by generating samples for 64 text prompts and measuring the mean CLIP similarity score. For a full list of prompts see Appendix B, some samples could not be run in DALLE-2 due to the content policy. Inference times are measured using and A100-80GB PCIe GPU, for generation of a single sample including CLIP re-ranking and do not take advantage of any batching. We note the DALLE-2-pytorch library does not batch the two calls to forward when using classifier free guidance and could be optimised further. Inference times are not included for DALLE-2 as it is not available to run on comparable hardware.}
    \label{table:compare}

    \begin{tabular}{lS[table-format=1.3]S[table-format=2.3]} 
     \toprule
     \textbf{Method} & \textbf{CLIP score} & \textbf{run-time (s)}  \\ [0.5ex] 
     \midrule
     clip2latent (ours)  & 0.316 & 11.760 \\
     clip2latent (ours) + timestep respacing   & 0.315 & 0.244 \\
     Direct Optimisation & 0.321 & 19.191  \\ 
     LAFITE & 0.278 & 0.045 \\ 
     DALLE-2 & 0.291{$^{*}$} & \textemdash \\
     \bottomrule
    \end{tabular}

\end{table}
    
Our method produces high quality, high resolution images based on text prompts without the need for paired image-text data during training. Figure \ref{fig:compare} shows examples of text-to-image generation via our method, compared against LAFITE and direct optimisation\cite{nshepperd1} of the $w$ vector (for further details of the baseline comparison methods see Appendix C). We judge the similarity of generated image and text prompts using the standard CLIP similarity score, and show that our approach acheives CLIP socres close to optimisation methods whilst largely avoiding image artefacts and performing inference more rapidly, see Table \ref{table:compare}.

\subsection{Importance of embedding augmentation}
\label{sec:noise}

Although CLIP encodes both images and text into a shared latent space it has been noted that the encodings for the two modalities are disjoint in this space\cite{hoppe1,Liang2022-cv}. As we wish to train a text to image model with no external data we are restricted to using the CLIP image embeddings of StyleGAN generated images. Although, previous works has shown that simply training models conditional on image embeddings and substituting text embeddings at inference time is sufficient to generate good quality results \cite{Wang2022-cf,Ramesh2022-ai}, we find the the addition of scaled Gaussian noise to the condition embedding vectors (as suggested in LAFITE\cite{Zhou2021-dl}) is crucial in allowing the model to generalise across both image and text embeddings produced by CLIP. 

Although the authors of LAFITE describe this as a method for producing "psuedo text encodings", we consider this to be more analogous to a method of data augmentation, noting that the "pseudo text" embeddings and original image embedding have a similarity score of around 0.8 (given a noise scaling of 0.75 using in LAFITE), far higher than any real text embedding with a well matched image embedding (typically in range 0.3-0.4\cite{hoppe1}). Without the addition of noise, the model learns to produce latent codes corresponding very closely to the input image, but fails to generalise to text embeddings. At very high noise levels the conditioning is less informative and the model performance deteriorates. We find the optimal level of noise scaling to be 1.0, see Appendix Figure 1.

\subsection{Making use of StyleGAN}
\label{sec:using}

\begin{figure}[htbp]
    \centering
    \includegraphics[width=\textwidth]{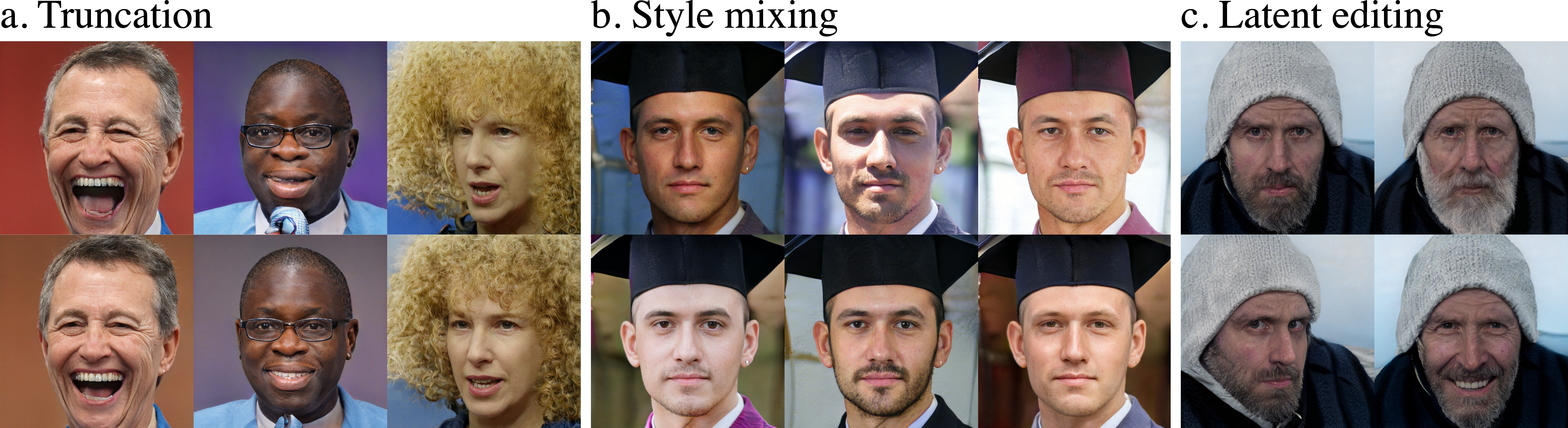}
    \caption{Demonstrating the application of existing StyleGAN editing techniques to add additional levels of control. a: Artefact correction using truncation, (top row original, bottom row truncation 0.8) for images of "a British politician laughing happily", "a Nigerian professor of economics", "person with very tight curly blonde hair". b: Style mixing to create variations of colour, lighting, and texture for "a university graduate". c: Age, Pose, and Smile editing using InterfaceGAN directions on a generated sample (top-left) from the prompt "an arctic explorer". }
    \label{fig:using}
\end{figure}

The image synthesis portion of our text-to-image pipeline uses the well studied and understood StyleGAN generator. This means we can leverage many of the well known and favourable properties of StyleGAN to exert further control over our generated images. In the following section we demonstrate how our method can take advantage of: truncation for artefact removal, style mixing for improving diversity, and the well-known editability of StyleGAN's latent space.

\subsubsection{Truncation}

We find that at high guidance scales our latent vectors can sometimes fall outside the typical domain of StyleGAN latent space, generating unnatural artefacts. One method to alleviate this is to use the well-known technique of truncation\cite{Brock2018-at} to move the generated latent closer to the mean latent vector. This trades off image-text similarity and diversity for image fidelity, see Figure \ref{fig:using}a. Although in some cases this can actually \textit{increase} the CLIP similarity score as reducing artefacts brings the image closer to the domain of real face image.

\subsubsection{Style Mixing}

We can also take advantage of the extended latent space of StyleGAN ($w+$) where the latent vector for each layer controls the generated image at a different resolution scale\cite{Karras2019-sj}. One straightforward way to utilise this structure is to add extra colour and lighting diversity to our generated images by performing Style Mixing with our generated latent vector for the lower resolution layers which control most of the semantic appearance of the generated image, with a randomly sampled latent for higher resolution layers which predominantly control colour and lighting, see Figure \ref{fig:using}b.

\subsubsection{Latent Editing}

Finally we can also leverage the extensive literature on finding semantically meaningful edits in StyleGAN latent space to edit our generated images, adding an extra level of control to our text-to-image pipeline. This allows us to make use of the wealth of existing high-quality facial editing directions available for StyleGAN. We show the application of some well known InterfaceGAN\cite{Shen2020-ur} directions to our generated images in Figure \ref{fig:using}c. It has previously been shown that high-quality edits require latent vectors to lie well within the typical domain of StyleGAN's latent space\cite{Tov2021-le}. The generative nature of the clip2latent diffusion model ensures that text-generated latents are within the domain of StyleGAN without the need for additional losses such as the "latent discriminator" in \textit{Tov et al.}\cite{Shen2020-ur}. This ensure that our method can generate high-quality edits with few artefacts, in contrast to alternatives such as direct optimisation which generates latents with poor editability, see Appendix Figure 2.

\section{Discussion}

We note our method is not restricted to StyleGAN in particular and in theory could be applied to any generative model. In practice our method appears to favour models with a relatively low dimensional latent space (see Appendix E) and clearly is well suited to those models for which high quality pre-trained networks are available. To demonstrate the generality of our method beyond faces we also train a text-to-image model using a 256x256 StyleGAN3 landscape model\cite{Pinkney_Awesome_Pretrained_StyleGAN3} trained on the LHQ dataset\cite{Skorokhodov2021-kw}, see Figure \ref{fig:headline}, bottom row. Throughout this work we have chosen the StyleGAN family of architectures due to the wide range of pre-trained models available\cite{Pinkney_Awesome_pretrained_StyleGAN2} and its high-resolution, fast inference and state of the art performance in many domains\cite{Sauer2022-bi,Kang2022-ax}. However, there is no reason a different GAN (e.g. BigGAN\cite{Brock2018-at}) or entirely different class of generative model (e.g. VAE\cite{Child2020-rx}) couldn't also be used.

Although we have chosen to take a simple data-free approach to training our model, which does not rely on any exist text-image paired data, there is now wide availability of such datasets across a variety of domains thanks to the open-source efforts of LAION\cite{Schuhmann2021-qv}.  To make use of such paired data would require embedding the real-world images into the latent space of StyleGAN for which there exist many methods\cite{Richardson2020-og,Abdal2019-nj,Alaluf2021-xu}. We leave this as an avenue for exploration for future work.

Our generated images are constrained to lie withing the $w$ space of StyleGAN, although this can produce a large variety of images its expensiveness is limited. Previous work has shown that the $w+$ space is vastly more capable in what images it can represent\cite{Abdal2019-nj}. We show some limited exploration in applying our model to other StyleGAN latent spaces in Appendix E, but leave extensions to the other StyleGAN latent spaces to future work.

As well as generalisation to other generative model latent spaces we believe that the application of powerful diffusion models can allow the addition of conditional sampling to previously unconditional models based on any image encoding, for example facial recognition/attribute networks or other classification models. We look forward to future applications of diffusion models as tools for arbitrarily mapping between latent spaces of pre-trained models. 

\section{Conclusion}

Our method allows sampling from a pre-trained generative model conditioned on CLIP image and text embeddings, providing a lightweight way to train and inference an effective text-to-image generation model without access to external data. Our method generates high quality results with satisfactory CLIP scores whilst largely avoiding image artefacts from earlier methods. We hope our work sheds a light on how the connection between text and image enables more powerful use of pre-trained generative models.

\bibliography{main}

\begin{thebibliography}{49}
\providecommand{\natexlab}[1]{#1}
\providecommand{\url}[1]{\texttt{#1}}
\expandafter\ifx\csname urlstyle\endcsname\relax
  \providecommand{\doi}[1]{doi: #1}\else
  \providecommand{\doi}{doi: \begingroup \urlstyle{rm}\Url}\fi

\bibitem[Abdal et~al.(2019)Abdal, Qin, and Wonka]{Abdal2019-nj}
Rameen Abdal, Yipeng Qin, and Peter Wonka.
\newblock {Image2StyleGAN}: How to embed images into the {StyleGAN} latent
  space?
\newblock April 2019.

\bibitem[Abdal et~al.(2021)Abdal, Zhu, Femiani, Mitra, and Wonka]{Abdal2021-lv}
Rameen Abdal, Peihao Zhu, John Femiani, Niloy~J Mitra, and Peter Wonka.
\newblock {CLIP2StyleGAN}: Unsupervised extraction of {StyleGAN} edit
  directions.
\newblock December 2021.

\bibitem[Alaluf et~al.(2021)Alaluf, Patashnik, and Cohen-Or]{Alaluf2021-xu}
Yuval Alaluf, Or~Patashnik, and Daniel Cohen-Or.
\newblock {ReStyle}: A {Residual-Based} {StyleGAN} encoder via iterative
  refinement.
\newblock April 2021.

\bibitem[Brock et~al.(2018)Brock, Donahue, and Simonyan]{Brock2018-at}
Andrew Brock, Jeff Donahue, and Karen Simonyan.
\newblock Large scale {GAN} training for high fidelity natural image synthesis.
\newblock September 2018.

\bibitem[Child(2020)]{Child2020-rx}
Rewon Child.
\newblock Very deep {VAEs} generalize autoregressive models and can outperform
  them on images.
\newblock November 2020.

\bibitem[Crowson(2021)]{crowson1}
Katherine Crowson.
\newblock Clip guided diffusion hq 256x256., 2021.
\newblock URL \url{https://colab.research.google.com/
  drive/12a_Wrfi2_gwwAuN3VvMTwVMz9TfqctNj}.

\bibitem[Ding et~al.(2022)Ding, Zheng, Hong, and Tang]{Ding2022-gr}
Ming Ding, Wendi Zheng, Wenyi Hong, and Jie Tang.
\newblock {CogView2}: Faster and better {Text-to-Image} generation via
  hierarchical transformers.
\newblock April 2022.

\bibitem[Gabbay et~al.(2021)Gabbay, Cohen, and Hoshen]{Gabbay2021-go}
Aviv Gabbay, Niv Cohen, and Yedid Hoshen.
\newblock An image is worth more than a thousand words: Towards disentanglement
  in the wild.
\newblock June 2021.

\bibitem[Gal et~al.(2021)Gal, Patashnik, Maron, Chechik, and
  Cohen-Or]{Gal2021-ir}
Rinon Gal, Or~Patashnik, Haggai Maron, Gal Chechik, and Daniel Cohen-Or.
\newblock {StyleGAN-NADA}: {CLIP-Guided} domain adaptation of image generators.
\newblock August 2021.

\bibitem[Gu et~al.(2021)Gu, Chen, Bao, Wen, Zhang, Chen, Yuan, and
  Guo]{Gu2021-tt}
Shuyang Gu, Dong Chen, Jianmin Bao, Fang Wen, Bo~Zhang, Dongdong Chen, Lu~Yuan,
  and Baining Guo.
\newblock Vector quantized diffusion model for {Text-to-Image} synthesis.
\newblock November 2021.

\bibitem[Ho and Salimans(2021)]{Ho2021-ck}
Jonathan Ho and Tim Salimans.
\newblock {Classifier-Free} diffusion guidance.
\newblock September 2021.

\bibitem[Ho et~al.(2020)Ho, Jain, and Abbeel]{Ho2020-dk}
Jonathan Ho, Ajay Jain, and Pieter Abbeel.
\newblock Denoising diffusion probabilistic models.
\newblock \emph{arXiv [cs.LG]}, June 2020.

\bibitem[Hoppe(2021)]{hoppe1}
Travis Hoppe.
\newblock Tweet, 2021.
\newblock URL
  \url{https://twitter.com/metasemantic/status/1356406256802607112}.

\bibitem[Kang et~al.(2022)Kang, Shin, and Park]{Kang2022-ax}
Minguk Kang, Joonghyuk Shin, and Jaesik Park.
\newblock {StudioGAN}: A taxonomy and benchmark of {GANs} for image synthesis.
\newblock June 2022.

\bibitem[Karras et~al.(2019)Karras, Laine, Aittala, Hellsten, Lehtinen, and
  Aila]{Karras2019-sj}
Tero Karras, Samuli Laine, Miika Aittala, Janne Hellsten, Jaakko Lehtinen, and
  Timo Aila.
\newblock Analyzing and improving the image quality of {StyleGAN}.
\newblock December 2019.

\bibitem[Karras et~al.(2021)Karras, Aittala, Laine, H{\"a}rk{\"o}nen, Hellsten,
  Lehtinen, and Aila]{Karras2021-sg}
Tero Karras, Miika Aittala, Samuli Laine, Erik H{\"a}rk{\"o}nen, Janne
  Hellsten, Jaakko Lehtinen, and Timo Aila.
\newblock Alias-free generative adversarial networks.
\newblock June 2021.

\bibitem[Karras et~al.(2022)Karras, Aittala, Aila, and Laine]{Karras2022-wf}
Tero Karras, Miika Aittala, Timo Aila, and Samuli Laine.
\newblock Elucidating the design space of diffusion-based generative models.
\newblock \emph{arXiv [cs.CV]}, June 2022.

\bibitem[Kocasari et~al.(2021)Kocasari, Dirik, Tiftikci, and
  Yanardag]{Kocasari2021-wv}
Umut Kocasari, Alara Dirik, Mert Tiftikci, and Pinar Yanardag.
\newblock {StyleMC}: {Multi-Channel} based fast {Text-Guided} image generation
  and manipulation.
\newblock December 2021.

\bibitem[Liang et~al.(2022)Liang, Zhang, Kwon, Yeung, and Zou]{Liang2022-cv}
Weixin Liang, Yuhui Zhang, Yongchan Kwon, Serena Yeung, and James Zou.
\newblock Mind the gap: Understanding the modality gap in multi-modal
  contrastive representation learning.
\newblock March 2022.

\bibitem[Murdock(2021{\natexlab{a}})]{murdock1}
Ryan Murdock.
\newblock Tweet, 2021{\natexlab{a}}.
\newblock URL \url{https://twitter.com/advadnoun/status/1348375026697834496}.

\bibitem[Murdock(2021{\natexlab{b}})]{murdock2}
Ryan Murdock.
\newblock Tweet, 2021{\natexlab{b}}.
\newblock URL \url{https://twitter.com/advadnoun/status/1351038053033406468}.

\bibitem[Nichol and Dhariwal(2021)]{Nichol2021-jd}
Alex Nichol and Prafulla Dhariwal.
\newblock Improved denoising diffusion probabilistic models.
\newblock February 2021.

\bibitem[Nichol et~al.(2021)Nichol, Dhariwal, Ramesh, Shyam, Mishkin, McGrew,
  Sutskever, and Chen]{Nichol2021-be}
Alex Nichol, Prafulla Dhariwal, Aditya Ramesh, Pranav Shyam, Pamela Mishkin,
  Bob McGrew, Ilya Sutskever, and Mark Chen.
\newblock {GLIDE}: Towards photorealistic image generation and editing with
  {Text-Guided} diffusion models.
\newblock December 2021.

\bibitem[nshepperd(2021)]{nshepperd1}
nshepperd.
\newblock stylegan3 with clip guidance, 2021.
\newblock URL
  \url{https://colab.research.google.com/drive/1eYlenR1GHPZXt-YuvXabzO9wfh9CWY36}.

\bibitem[Olah et~al.(2017)Olah, Mordvintsev, and Schubert]{Olah2017-lu}
Chris Olah, Alexander Mordvintsev, and Ludwig Schubert.
\newblock Feature visualization.
\newblock \emph{Distill}, 2\penalty0 (11), November 2017.

\bibitem[Patashnik et~al.(2021)Patashnik, Wu, Shechtman, Cohen-Or, and
  Lischinski]{Patashnik2021-zm}
Or~Patashnik, Zongze Wu, Eli Shechtman, Daniel Cohen-Or, and Dani Lischinski.
\newblock {StyleCLIP}: {Text-Driven} manipulation of {StyleGAN} imagery.
\newblock March 2021.

\bibitem[Pinkney({\natexlab{a}})]{Pinkney_Awesome_Pretrained_StyleGAN3}
Justin N.~M. Pinkney.
\newblock {Awesome Pretrained StyleGAN3}, {\natexlab{a}}.

\bibitem[Pinkney({\natexlab{b}})]{Pinkney_Awesome_pretrained_StyleGAN2}
Justin N.~M. Pinkney.
\newblock {Awesome pretrained StyleGAN2}, {\natexlab{b}}.

\bibitem[Radford et~al.(2021)Radford, Kim, Hallacy, Ramesh, Goh, Agarwal,
  Sastry, Askell, Mishkin, Clark, Krueger, and Sutskever]{Radford2021-ru}
Alec Radford, Jong~Wook Kim, Chris Hallacy, Aditya Ramesh, Gabriel Goh,
  Sandhini Agarwal, Girish Sastry, Amanda Askell, Pamela Mishkin, Jack Clark,
  Gretchen Krueger, and Ilya Sutskever.
\newblock Learning transferable visual models from natural language
  supervision.
\newblock February 2021.

\bibitem[Ramesh et~al.(2021)Ramesh, Pavlov, Goh, Gray, Voss, Radford, Chen, and
  Sutskever]{Ramesh2021-ro}
Aditya Ramesh, Mikhail Pavlov, Gabriel Goh, Scott Gray, Chelsea Voss, Alec
  Radford, Mark Chen, and Ilya Sutskever.
\newblock {Zero-Shot} {Text-to-Image} generation.
\newblock February 2021.

\bibitem[Ramesh et~al.(2022)Ramesh, Dhariwal, Nichol, Chu, and
  Chen]{Ramesh2022-ai}
Aditya Ramesh, Prafulla Dhariwal, Alex Nichol, Casey Chu, and Mark Chen.
\newblock Hierarchical {Text-Conditional} image generation with {CLIP} latents.
\newblock April 2022.

\bibitem[Richardson et~al.(2020)Richardson, Alaluf, Patashnik, Nitzan, Azar,
  Shapiro, and Cohen-Or]{Richardson2020-og}
Elad Richardson, Yuval Alaluf, Or~Patashnik, Yotam Nitzan, Yaniv Azar, Stav
  Shapiro, and Daniel Cohen-Or.
\newblock Encoding in style: a {StyleGAN} encoder for {Image-to-Image}
  translation.
\newblock August 2020.

\bibitem[Rombach et~al.(2021)Rombach, Blattmann, Lorenz, Esser, and
  Ommer]{Rombach2021-bu}
Robin Rombach, Andreas Blattmann, Dominik Lorenz, Patrick Esser, and Bj{\"o}rn
  Ommer.
\newblock {High-Resolution} image synthesis with latent diffusion models.
\newblock December 2021.

\bibitem[Saharia et~al.(2022)Saharia, Chan, Saxena, Li, Whang, Denton,
  Ghasemipour, Ayan, Mahdavi, Lopes, Salimans, Ho, Fleet, and
  Norouzi]{Saharia2022-ry}
Chitwan Saharia, William Chan, Saurabh Saxena, Lala Li, Jay Whang, Emily
  Denton, Seyed Kamyar~Seyed Ghasemipour, Burcu~Karagol Ayan, S~Sara Mahdavi,
  Rapha~Gontijo Lopes, Tim Salimans, Jonathan Ho, David~J Fleet, and Mohammad
  Norouzi.
\newblock Photorealistic text-to-image diffusion models with deep language
  understanding.
\newblock \emph{arXiv [cs.CV]}, May 2022.

\bibitem[Sauer et~al.(2022)Sauer, Schwarz, and Geiger]{Sauer2022-bi}
Axel Sauer, Katja Schwarz, and Andreas Geiger.
\newblock {StyleGAN-XL}: Scaling {StyleGAN} to large diverse datasets.
\newblock February 2022.

\bibitem[Schuhmann et~al.(2021)Schuhmann, Vencu, Beaumont, Kaczmarczyk, Mullis,
  Katta, Coombes, Jitsev, and Komatsuzaki]{Schuhmann2021-qv}
Christoph Schuhmann, Richard Vencu, Romain Beaumont, Robert Kaczmarczyk,
  Clayton Mullis, Aarush Katta, Theo Coombes, Jenia Jitsev, and Aran
  Komatsuzaki.
\newblock {LAION-400M}: Open dataset of {CLIP-Filtered} 400 million
  {Image-Text} pairs.
\newblock November 2021.

\bibitem[Shen et~al.(2020)Shen, Yang, Tang, and Zhou]{Shen2020-ur}
Yujun Shen, Ceyuan Yang, Xiaoou Tang, and Bolei Zhou.
\newblock {InterFaceGAN}: Interpreting the disentangled face representation
  learned by {GANs}.
\newblock May 2020.

\bibitem[Skorokhodov et~al.(2021)Skorokhodov, Sotnikov, and
  Elhoseiny]{Skorokhodov2021-kw}
Ivan Skorokhodov, Grigorii Sotnikov, and Mohamed Elhoseiny.
\newblock Aligning latent and image spaces to connect the unconnectable.
\newblock April 2021.

\bibitem[Sohl-Dickstein et~al.(2015)Sohl-Dickstein, Weiss, Maheswaranathan, and
  Ganguli]{Sohl-Dickstein2015-hz}
Jascha Sohl-Dickstein, Eric~A Weiss, Niru Maheswaranathan, and Surya Ganguli.
\newblock Deep unsupervised learning using nonequilibrium thermodynamics.
\newblock March 2015.

\bibitem[Song and Ermon(2020)]{Song2020-cd}
Yang Song and Stefano Ermon.
\newblock Improved techniques for training {Score-Based} generative models.
\newblock June 2020.

\bibitem[Tov et~al.(2021)Tov, Alaluf, Nitzan, Patashnik, and
  Cohen-Or]{Tov2021-le}
Omer Tov, Yuval Alaluf, Yotam Nitzan, Or~Patashnik, and Daniel Cohen-Or.
\newblock Designing an encoder for {StyleGAN} image manipulation.
\newblock February 2021.

\bibitem[Tzelepis et~al.(2022)Tzelepis, Oldfield, Tzimiropoulos, and
  Patras]{Tzelepis2022-fj}
Christos Tzelepis, James Oldfield, Georgios Tzimiropoulos, and Ioannis Patras.
\newblock {ContraCLIP}: Interpretable {GAN} generation driven by pairs of
  contrasting sentences.
\newblock June 2022.

\bibitem[Wang(2022)]{wang1}
Phil Wang.
\newblock Dall-e 2 - pytorch, 2022.
\newblock URL \url{https://github.com/lucidrains/DALLE2-pytorch}.

\bibitem[Wang et~al.(2022{\natexlab{a}})Wang, Zhang, Zhang, Ouyang, Chen, Chen,
  and Wen]{Wang2022-ic}
Tengfei Wang, Ting Zhang, Bo~Zhang, Hao Ouyang, Dong Chen, Qifeng Chen, and
  Fang Wen.
\newblock Pretraining is all you need for {Image-to-Image} translation.
\newblock May 2022{\natexlab{a}}.

\bibitem[Wang et~al.(2022{\natexlab{b}})Wang, Liu, He, Wu, and Yi]{Wang2022-cf}
Zihao Wang, Wei Liu, Qian He, Xinglong Wu, and Zili Yi.
\newblock {CLIP-GEN}: {Language-Free} training of a {Text-to-Image} generator
  with {CLIP}.
\newblock March 2022{\natexlab{b}}.

\bibitem[Xu et~al.(2021)Xu, Lin, Tang, Li, He, Sebe, Timofte, Van~Gool, and
  Ding]{Xu2021-ev}
Zipeng Xu, Tianwei Lin, Hao Tang, Fu~Li, Dongliang He, Nicu Sebe, Radu Timofte,
  Luc Van~Gool, and Errui Ding.
\newblock Predict, prevent, and evaluate: Disentangled {Text-Driven} image
  manipulation empowered by {Pre-Trained} {Vision-Language} model.
\newblock November 2021.

\bibitem[Zhou et~al.(2021)Zhou, Zhang, Chen, Li, Tensmeyer, Yu, Gu, Xu, and
  Sun]{Zhou2021-dl}
Yufan Zhou, Ruiyi Zhang, Changyou Chen, Chunyuan Li, Chris Tensmeyer, Tong Yu,
  Jiuxiang Gu, Jinhui Xu, and Tong Sun.
\newblock {LAFITE}: Towards {Language-Free} training for {Text-to-Image}
  generation.
\newblock November 2021.

\bibitem[Zhu et~al.(2020)Zhu, Abdal, Qin, Femiani, and Wonka]{Zhu2020-bj}
Peihao Zhu, Rameen Abdal, Yipeng Qin, John Femiani, and Peter Wonka.
\newblock Improved {StyleGAN} embedding: Where are the good latents?
\newblock December 2020.

\bibitem[Zhu et~al.(2021)Zhu, Abdal, Femiani, and Wonka]{Zhu2021-bw}
Peihao Zhu, Rameen Abdal, John Femiani, and Peter Wonka.
\newblock Mind the gap: Domain gap control for single shot domain adaptation
  for generative adversarial networks.
\newblock October 2021.

\end{thebibliography}

\appendix
\appendixpage

\section{Hyperparameters}
\label{app:model}

\subsection{Noise level}
\label{app:noise}

\begin{figure}[htbp]
    \centering
    \includegraphics[width=0.5\textwidth]{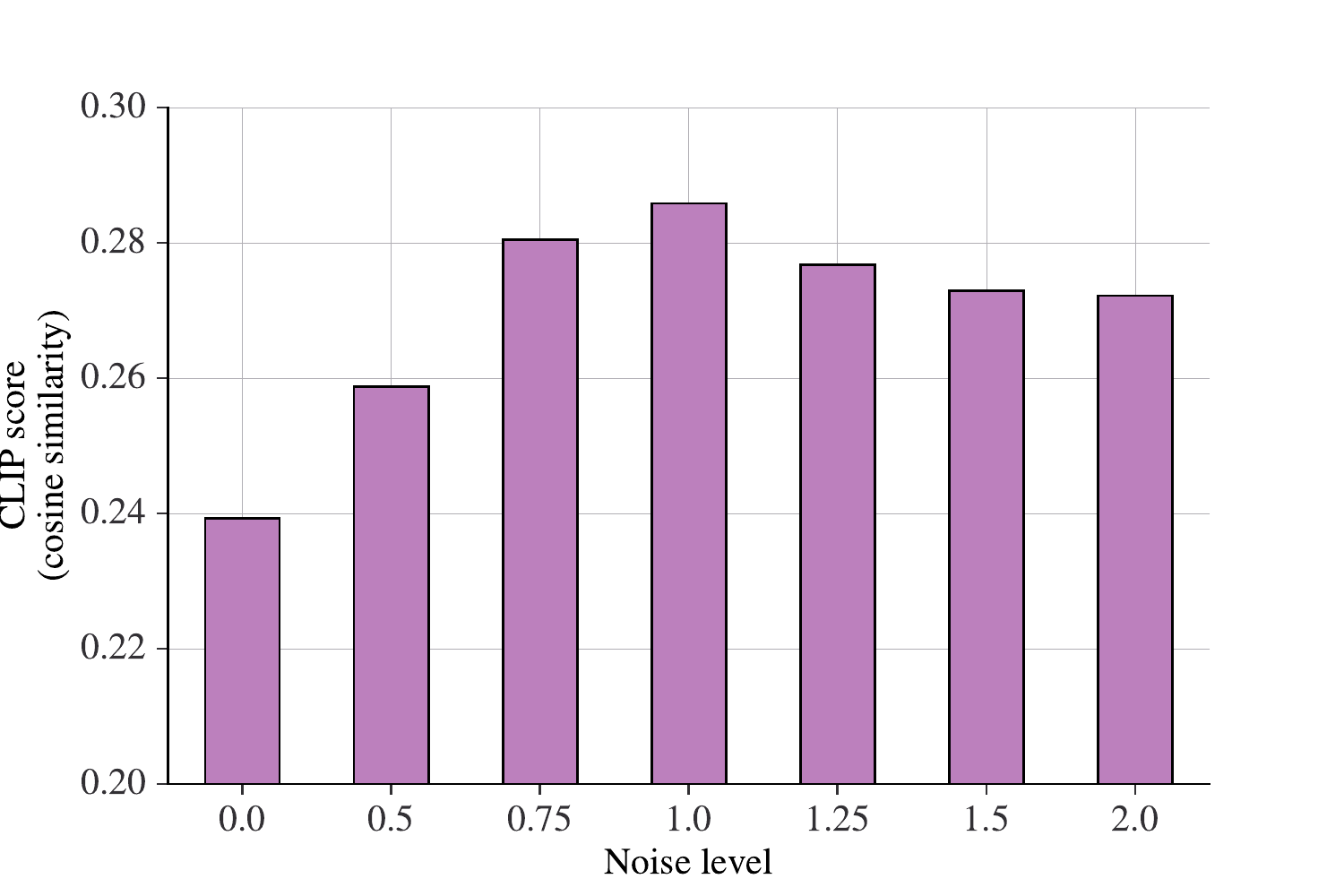}
    \caption{CLIP similarity score measured at different noise scaling factors ($\alpha$) applied to the CLIP image embeddings on which the model is conditioned.}
    \label{fig:noise}
\end{figure}

Following LAFITE\cite{Zhou2021-dl} we apply scaled Gaussian noise to generate an augmented CLIP embeddings, $e_i^\prime$, as follows:

\begin{equation}
e_i^\prime = \frac{y}{||y||_2}, \quad y = e_i + \alpha \frac{\epsilon}{||\epsilon||_2}
\end{equation}

where $e_i$ represents the normalised CLIP image embedding, $\epsilon \sim \mathcal{N}(0,I)$ is a Gaussian noise sample of the same dimension as $e_i$, and $\alpha$ is the noise scaling factor.

To ascertain the optimal noise level for CLIP image embedding augmentations we perform a simple search of possible noise values. For each noise level we run training of clip2latent model using the StyleGAN2 FFHQ model for 100 million iterations. We then select the best model by generating images using a set of 16 text prompts and measuring the CLIP similarity score. We find that the optimal value for $\alpha$ is $1.0$, see Figure \ref{fig:noise}

\subsection{Model hyper parameters}

Model and training hyper-parameters used for training both the StyleGAN2 FFHQ and StyleGAN3 LHQ clip2latent models. Parameter names follow the keys in the training configurations available in the code repository at: \url{https://github.com/justinpinkney/clip2latent}. In total our model has 48.9 million parameters.

\begin{table}[h]
    \centering
    \begin{tabular}{ cc }
        Model parameters & Training parameters \\
        \begin{tabular}{rc}
        \toprule
        \textbf{Parameter} & \textbf{Value} \\ \midrule
        timesteps          & 1000           \\ 
        beta\_schedule     & cosine         \\ 
        predict\_x\_start  & True           \\ 
        cond\_drop\_prob   & 0.2            \\ 
        dim                & 512            \\ 
        depth              & 12             \\ 
        dim\_head          & 64             \\ 
        heads              & 12             \\ \bottomrule
        \end{tabular}
        &
        \begin{tabular}{rc}
        \toprule
        \textbf{Parameter} & \textbf{Value} \\ \midrule
        iterations          & 1,000,000           \\ 
        batch\_size     & 512         \\ 
        lr  & 1.0e-4           \\ 
        weight\_decay   & 1.0e-2            \\ 
        ema\_beta                & 0.9999            \\ 
        ema\_update\_every              & 10             \\ 
        noise\_scale ($\alpha$)         & 1.0             \\
        Adam $\beta _1$, $\beta _2$ & 0.9, 0.999 \\ \bottomrule
        \end{tabular}
    \end{tabular}
    \label{table:hparams}

\end{table}

\section{CLIP score prompts}
\label{app:prompts}
Below is the list of prompts used for CLIP similarity scoring with clip2latent and comparable methods, all captiosn were prefixed by "A photograph of". An asterisk indicates the prompt could not be used with DALLE-2 due to the content policy applied by OpenAI.

\tiny
\begin{multicols}{3}
\begin{enumerate}
\itemsep-1em 
\parsep-1em
\item a person with glasses
\item a person with brown hair
\item a person with curly blonde hair
\item a person with a hat
\item a person with bushy eyebrows and a small mouth
\item a person smiling
\item a person who is angry
\item a person looking up at the sky
\item a person with their eyes closed
\item a person talking
\item a man with a beard
\item a happy man with a moustache
\item a young man
\item an old man
\item a middle aged man
\item a youthful man with a bored expression
\item a woman with a hat
\item a happy woman with glasses
\item a young woman
\item an old woman
\item a middle aged woman
\item a baby crying in a red bouncer
\item a child with blue eyes and straight brown hair in the sunshine
\item an old woman with large sunglasses and ear rings
\item a young man with a bald head who is wearing necklace in the city at night
\item a youthful woman with a bored expression
\item President Xi Jinping *
\item Prime Minister Boris Johnson *
\item President Joe Biden *
\item President Barack Obama *
\item Chancellor Angela Merkel *
\item President Emmanuel Macron *
\item Prime Minister Shinzo Abe *
\item Robert De Niro
\item Danny Devito
\item Denzel Washington
\item Meryl Streep *
\item Cate Blanchett *
\item Morgan Freeman *
\item Whoopi Goldberg *
\item Usain Bolt *
\item Muhammad Ali *
\item Serena Williams *
\item Roger Federer *
\item Martina Navratilova *
\item Jessica Ennis-Hill *
\item Cathy Freeman *
\item Christiano Ronaldo *
\item Elsa from Frozen
\item Eric Cartman from South Park
\item Chihiro from Spirited Away
\item Bart from the Simpsons
\item Woody from Toy Story
\item a university graduate
\item a firefighter
\item a police officer
\item a butcher
\item a scientist
\item a gardener
\item a hairdresser
\item a man visiting the beach
\item a woman giving a TED talk
\item a child playing with friends
\item a person watching birds in the forest
\end{enumerate}
\end{multicols}
\normalsize

\section{Comparison methods}
\label{app:compare}

\subsection{Direct latent optimisation}

We employ code by nshepperd\cite{nshepperd1} for optimisation of StyleGAN latent vectors based on CLIP similarity score to a text prompt. We modify the existing code such that optimisation is performed in $w$ space (rather than the original $w+$ space) of StyleGAN to more closely match our approach. We otherwise leave optimisation parameters as originally shared in the notebook.

\subsection{LAFITE text-free model}

We use the text-free MM-CelebA trained model provided by the original authors of LAFITE. As the inference of LAFITE is fast, to most closely compare LAFITE with our method we employ the same CLIP re-ranking approach as clip2latent, and sample 16 generations for each prompt and select that with the best CLIP similiary score.

\subsection{DALLE-2}

We compare to DALLE-2 using the web interface provided by OpenAI. The web interface was returns 4 images per prompt, to avoid bias the first sample was taken for every prompt. To try and generate images more comparable with the style of other methods trial and error was used to generate an engineered prompt of the form: "A photo headshot, single whole head centred, of $\langle caption \rangle$, flickr dslr, portrait photograph"
\section{Latent Editability}

\begin{figure}[htbp]
    \centering
    \includegraphics[width=\textwidth]{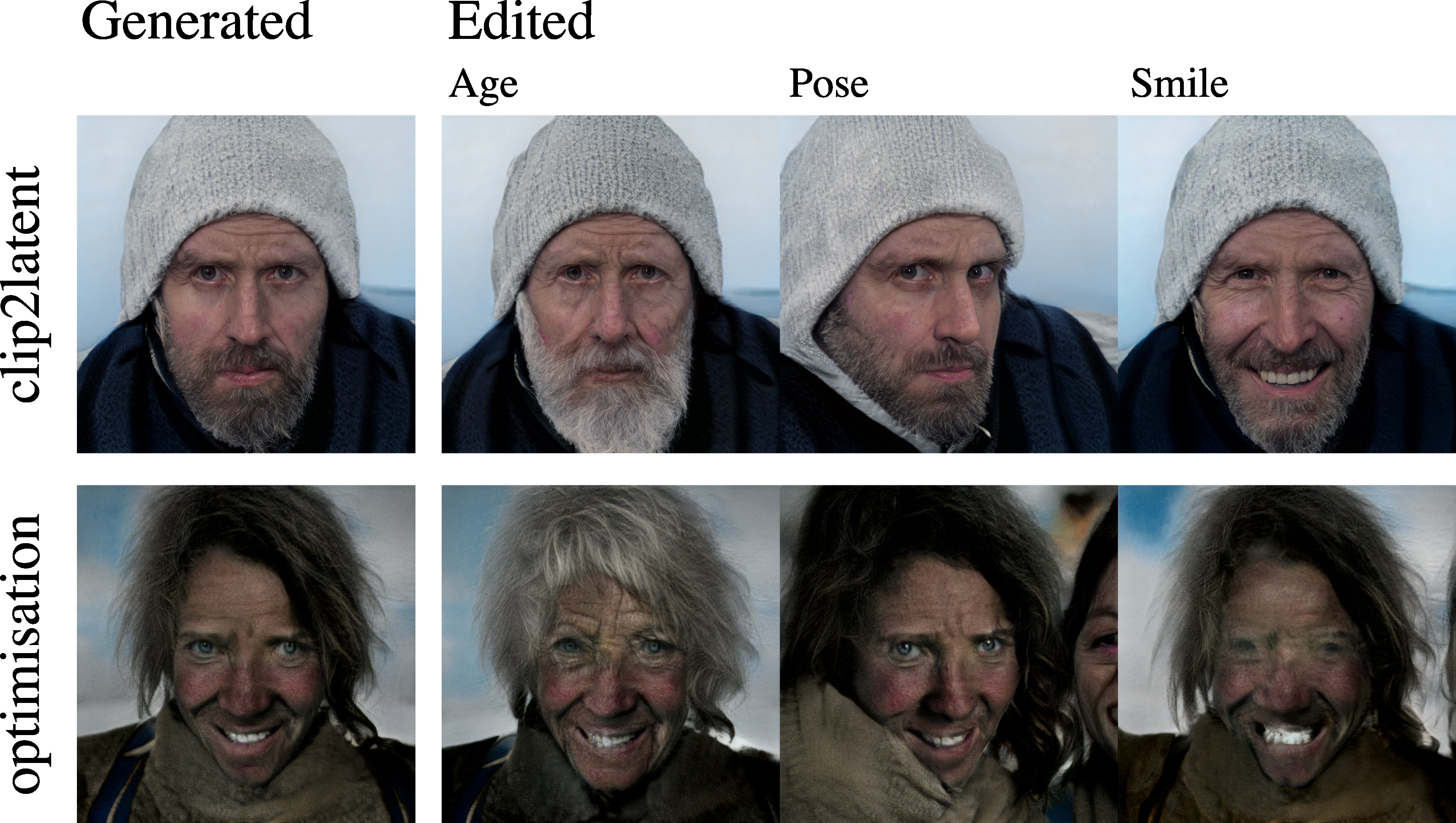}
    \caption{Comparison of editability of latents generated from the text prompt "A photograph of an arctic explorer" generated using clip2latent vs direct optimisation. Edit directions were applied with the same magnitude to both generated latents.}
    \label{fig:opt-edit}
\end{figure}

Our methods generate latent vectors which can effectively be edit using existing editing directions in StyleGAN's latent space. In contrast edits of the same magnitude applied to latents produced by direct optimisation frequently produce artefacts such as changes in composition, poor image quality, and in some cases complete failure to generate a recognisable face, see Figure \ref{fig:opt-edit}.

\section{Other StyleGAN latent spaces}
\label{app:w3}

We also explored the training of clip2latent model using the extended latent spaces of StyleGAN. The $w+$ space has been shown to be more expressive then $w$ and is frequently used to accurately embed arbitrary images into the StyleGAN latent space\cite{Abdal2019-nj}. However generating samples from $w+$ by randomly combining $w$ samples in from all 18 vectors gives rise to low quality generated images\cite{Zhu2020-bj} making it unsuitable for our application.

We briefly explore the use of a more limited latent space comprising three independent $w$ vectors which correspond to low, medium and high resolution layers. We term this latent space $w3$ and find that randomly generated samples are still high-quality but more diverse. We explored training a clip2latent model in this space using the same hyper-parameters as in $w$ space. We found that although the model trains successfully, the overall CLIP similarity scores are not improved and the model converges more slowly. We leave further exploration of how to exploit the greater diversity of extended latent spaces to future work.

\section{Finding edit directions}
\label{app:direction}

\begin{figure}[htbp]
    \centering
    \includegraphics[width=\textwidth]{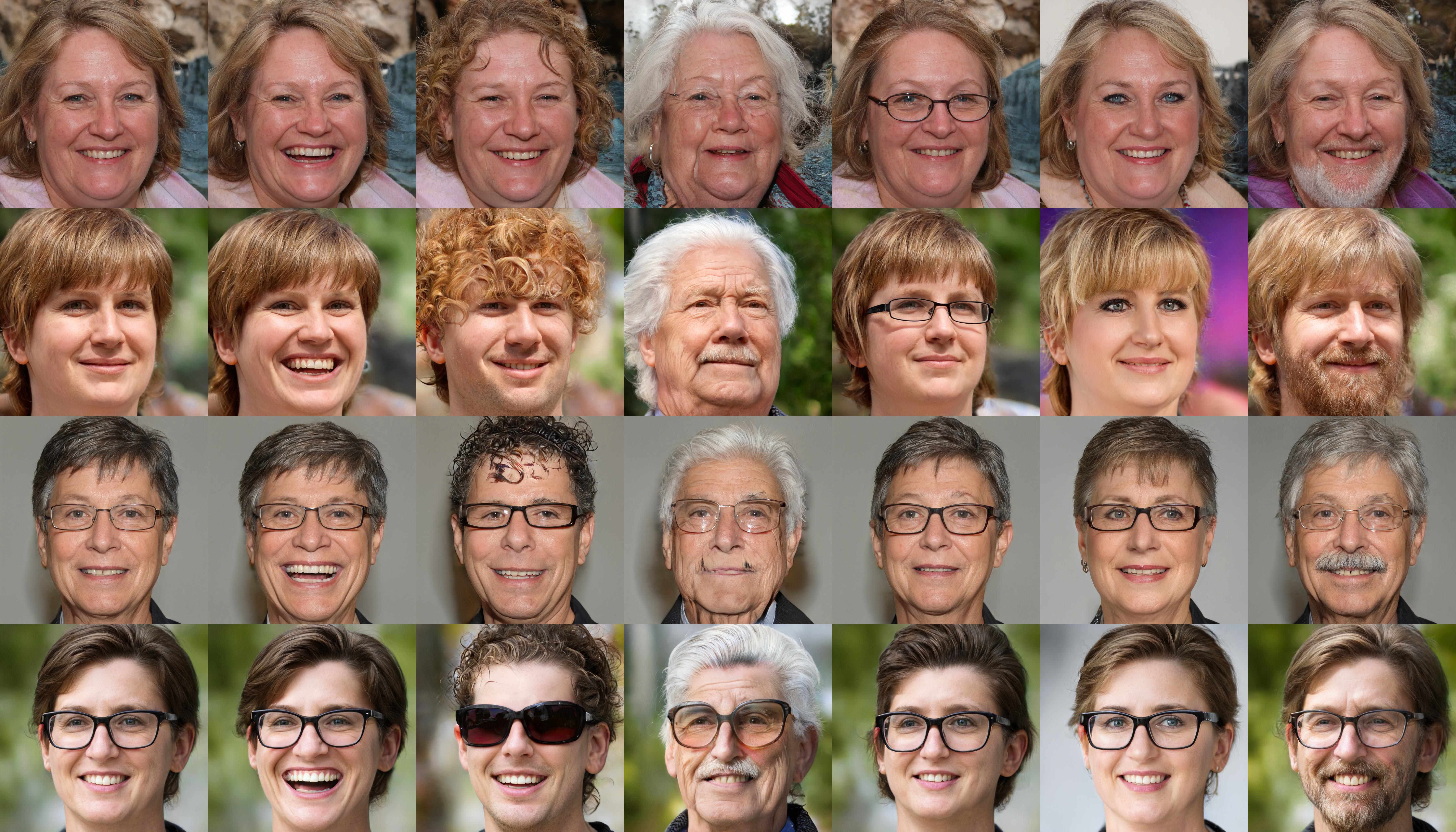}
    \caption{Examples of editing direction found using text based latent vector generation using clip2latent. Left column: random samples from StyleGAN. Following columns: edits in the directions: smile, curly hair, age, glasses, make up, beard.}
    \label{fig:edit-find}
\end{figure}

ALthough not our main application we also note that our latent generation method can be used to find directions within StyleGAN's latent space. By generating sets of latent vectors corresponding to a "positive" and "negative" prompt we can measure the direction between the two sets in order to obtain a latent direction, see Figure \ref{fig:edit-find}.

\end{document}